\documentclass[sigconf]{acmart}
\AtBeginDocument{%
  \providecommand\BibTeX{{%
    \normalfont B\kern-0.5em{\scshape i\kern-0.25em b}\kern-0.8em\TeX}}}

\setcopyright{none}
\copyrightyear{2022}
\acmYear{2022}
\acmDOI{XXXXXXX.XXXXXXX}
\renewcommand\footnotetextcopyrightpermission[1]{}
\usepackage{multirow}

\begin{document}
%% 一些问题
%% 一张图好像说不清我的模型流程，目前一张图用来说训练流程，但缺少说明结构的
%% 消融实验要补充

\title{Semi-Supervised Clustering with Contrastive Learning for Discovering New Intents}


\author{Feng Wei}
\authornote{Both authors contributed equally to this research.}
\email{huodeng.wf@antgroup.com}
\author{Zhenbo Chen}
\authornotemark[1]
\email{chenzhenbo.czb@alibaba-inc.com}
\affiliation{%
  \institution{MYbank, Ant Group}}

\author{Zhenghong Hao}
\email{haozhenghong.hzh@mybank.cn}
\affiliation{%
  \institution{MYbank, Ant Group}}
  
\author{Fengxin Yang}
\email{yangfengxin.yfx@alibaba-inc.com}
\affiliation{%
  \institution{MYbank, Ant Group}}
  
\author{Hua Wei}
\email{shuhu.wh@antgroup.com}
\affiliation{%
  \institution{MYbank, Ant Group}}
 
\author{Bing Han}
\email{hanbing.hanbing@antgroup.com}
\affiliation{%
  \institution{MYbank, Ant Group}}
 
\author{Sheng Guo}
\authornote{Corresponding author.}
\email{guosheng.guosheng@alibaba-inc.com}
\affiliation{%
  \institution{MYbank, Ant Group}}

\renewcommand{\shortauthors}{ }

\begin{abstract}
Most dialogue systems in real world rely on predefined intents and answers for QA service, so discovering potential intents from large corpus previously is really important for building such dialogue services. Considering that most scenarios have few intents known already and most intents waiting to be discovered, we focus on semi-supervised text clustering and try to make the proposed method benefit from labeled samples for better overall clustering performance. In this paper, we propose Deep Contrastive Semi-supervised Clustering (DCSC), which aims to cluster text samples in a semi-supervised way and provide grouped intents to operation staff. To make DCSC fully utilize the limited known intents, we propose a two-stage training procedure for DCSC, in which DCSC will be trained on both labeled samples and unlabeled samples, and achieve better text representation and clustering performance. We conduct experiments on two public datasets to compare our model with several popular methods, and the results show DCSC achieve best performance across all datasets and circumstances, indicating the effect of the improvements in our work.

\end{abstract}





\maketitle

\section{Introduction}
In real applications, many task-oriented dialogue systems are mostly based on Natural Language Understanding (NLU) to classify or match a user query into a known category and reply with a prepared answer. If we can discover as much new intents as possible, then chat robots will be able to answer many kinds of questions and will improve the user experience. To discover the intents, we need to group different samples with similar intents together through clustering techniques, and every cluster will be treated as a new potential intent. Meanwhile, the accuracy of clustering also matters for NLU modules, because clustered samples will be used for training a classification model or building distance-matching model. If a cluster contains too much noisy samples, the downstream NLU module may not recognize user intents correctly. Therefore, a well-performing chat robot depends on not only NLU abilities, but also some preparatory works like intent clustering.

Since intent discovery is critical for chat robots nowadays, there have been lots of works proposed in this specific field or in related fields. Early works mainly focus on unsupervised clustering, in which all samples will be treated as unlabeled for clustering. The most basic method for unsupervised clustering is the combination of a encoder model and a clustering model. In Natural Language Processing (NLP) tasks, the feature-extracting encoder can be language model such as BERT\cite{devlin2018bert} and SBERT\cite{reimers2019sentence}, and the clustering model can be machine learning methods such as K-Means++\cite{arthur2006k} and HDBSCAN\cite{campello2013density}. However, such methods separate the encoding step and clustering step, which cannot optimize the representation according to the clustering loss. To solve this problem, some early works use deep-learning-based clustering methods, such as DEC\cite{xie2016unsupervised} and DCN\cite{yang2017towards}, which associate representation and unsupervised clustering as a simultaneous optimization procedure and improve the final performance. In more recent researches, contrastive learning has been introduced to further improve the representations. In DeepCluster\cite{caron2018deep} and SwAV\cite{caron2020unsupervised}, contrastive learning as well as deep-learning-based clustering, greatly improve the representations of images for downstream tasks. In SCCL\cite{zhang2021supporting}, improved DEC\cite{xie2016unsupervised} with contrastive learning, has achieved ideal clustering performance for unsupervised text clustering.

However, in common scenarios, there will be few labelled samples of limited known intents available, and quite a lot of raw corpus waiting to be classified into known or unknown intents. Take our experience for example, when we are going to build a task-oriented chat robot, we will borrow some labeled corpus from another task (which contains some intents in common across different tasks), and will try to supplement new intents continuously. Unsupervised methods cannot benefit from these labeled samples and further improve the performance, therefore recently some researches have been work on semi-supervised models to utilize the limited supervised information. CDAC+\cite{lin2020discovering} uses labeled samples for pairwise similarities to guide the clustering process. DeepAligned\cite{zhang2021discovering} trains a better encoder through classification loss on labeled samples, and then iteratively train the encoder through pseudo labels produced by K-Means, which previously has achieved state-of-the-art results. Although these methods successfully utilize known intents, we think there is still space for improvements (for example, training of DeepAligned\cite{zhang2021discovering} lacks distance constrain which is more friendly for clustering, and this method still relies on K-Means for updating pseudo labels which is not robust as deep-learning-based methods).

In summary, there are two ways for improving the clustering, the first is utilizing labeled intents for better initial text representations, the second is building deep learning model for joint optimization for both representation and clustering. To solve these two problems, we propose Deep Contrastive Semi-supervised Clustering (DCSC). DCSC bases on BERT\cite{devlin2018bert} as backbone, and it is trained through a two-stage dual-task process. In stage one, DCSC is trained on labeled samples through Cross Entropy Loss and Supervised Contrastive Loss\cite{khosla2020supervised} for distance constrain, and is trained on unlabeled samples through Contrastive Loss as well. In stage two, we build a classifier head for to produce pseudo labels, and train DCSC using Cross Entropy Loss and Supervised Contrastive Loss\cite{khosla2020supervised} on samples with either ground truth labels or pseudo labels. We conduct experiments on two public datasets, Clinc\cite{larson2019evaluation} and Banking\cite{casanueva2020efficient}. To simulate the real situation and make our experiments comparable with previous works, we keep the same experiment settings as DeepAligned\cite{zhang2021discovering}, and evaluate models using Accuracy (ACC), Adjusted Rand Index (ARI), Normalized Mutual Information (NMI). Our experiments show DCSC has outperformed other methods with $10\%$ advantage at most under different experiment settings. Therefore, our contributions can be summarised as:

\begin{itemize}

\item Applying Unsupervised Contrastive Loss and Supervised Contrastive Loss\cite{khosla2020supervised} for unlabeled samples and labeled samples, which makes the text representation from backbone more friendly for clustering task.

\item Building a deep-learning-based clustering approach for semi-supervised tasks, which jointly optimize the clustering ability and representation ability.

\item Conducting comparative experiments on two public datasets, demonstrating that DCSC is well-performing and robust for text clustering under different experiment circumstances.

\end{itemize}

The rest of this paper is organized as follows. In Section 2, we introduce some previous related work which have inspired us. In Section 3, we discuss our proposed approach as a semi-supervised text clustering model. Then we conclude our experiments on public datasets in Section 4, and make the conclusion in Section 5.

\section{Related Work}
Our model DCSC mainly utilizes contrastive learning and deep clustering to achieve current performance. In this section, we are going to discuss some previous works that has inspired us from contrastive learning, deep clustering, and semi-supervised clustering.

\textbf{Contrastive Learning.}
To optimize representation in unsupervised way, contrastive learning augment samples for different views, and train the model to distinguish views of the same sample from a large batch. As shown SimCLR\cite{chen2020simple}, model pretrained with contrastive learning achieves accurate performance in downstream tasks. Meanwhile, SimCSE\cite{gao2021simcse} provides a simple but effective idea for data augmentation of NLP tasks when applying contrastive learning, it lets a sample propagate through backbone with dropout twice to get different embeddings and conduct contrastive learning on such outputs, and it achieves an average of $76.3\%$ Spearman’s correlation respectively with BERT\cite{devlin2018bert} (base)
on standard semantic textual similarity (STS) tasks. Also, contrastive learning can also be extended to supervised tasks\cite{khosla2020supervised}, trying to pull the samples belonging to the same class together in embedding space, while push apart samples from different classes.

\textbf{Deep Clustering.}
Jointly optimizing representation and clustering through deep networks, will guide the representation to be more suitable for clustering space. Early works like DEC\cite{xie2016unsupervised} replace K-Means with deep networks and iteratively optimize networks. However, such deep learning methods may cause trivial solution because most instances might be assigned to the a single cluster. SwAV\cite{caron2020unsupervised} solve this as a optimal transport problem, it uses Sinkhorn-Knopp algorithm\cite{cuturi2013sinkhorn} to produce soft pseudo assignments and optimize the networks through backpropagation. Meanwhile, similar to contrastive learning, it uses a “swapped” prediction mechanism where the model is trained to make prediction for a view under the soft assignment from another view's representation, trying to align the representation of different views. Such training strategy make SwAV\cite{caron2020unsupervised} achieve $75.3\%$ top-1 accuracy on ImageNet with ResNet-50.

\textbf{Semi-Supervised Clustering.}
Recently, some researches has contributed to intent discovering with semi-supervised clustering. DeepAligned\cite{zhang2021discovering} proposes a two-stage training strategy, in which the backbone is firstly trained to classify labeled samples for better representation (supervised learning) , and then secondly trained to classify samples with pseudo labels produced by K-Means iteratively. This paper has also conducted experiments on Clinc\cite{larson2019evaluation} and Banking\cite{casanueva2020efficient}, and randomly chooses a fraction of intents as known ones. DeepAligned\cite{zhang2021discovering} has outperformed other methods across all experiment settings and become the state-of-the-art model at the time. Also, with the same experiment settings, SCL\cite{shen2021semi} has achieved better results recently mainly with the improvements from contrastive learning and better backbone (MPNet\cite{song2020mpnet}).

\section{Our Approach}
\begin{figure*}[ht]
  \centering
  \includegraphics[width=\textwidth]{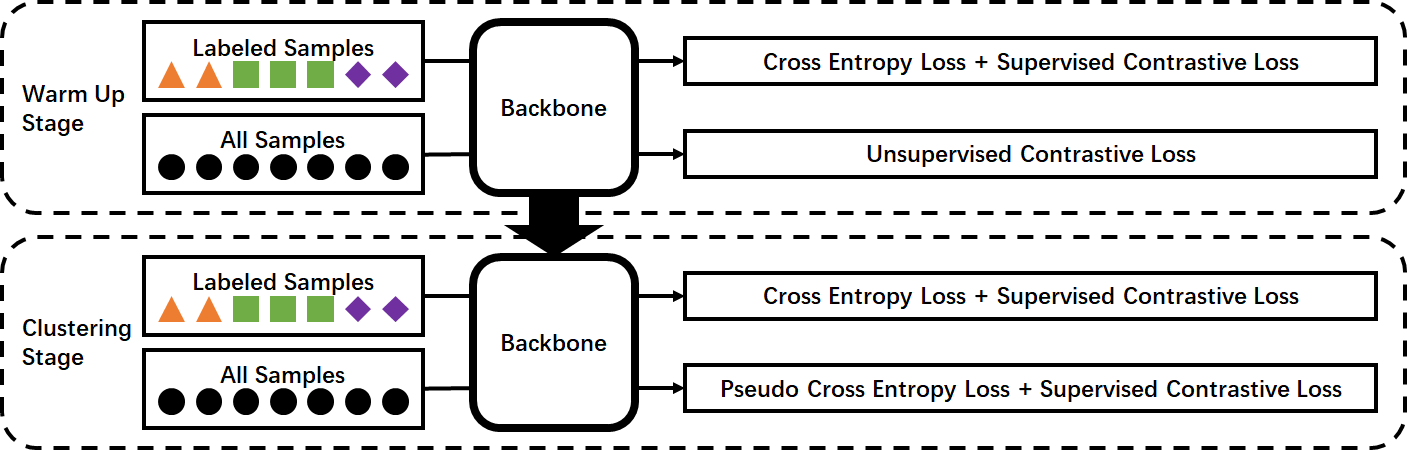}
  \caption{Training procedure of DCSC. In warm up stage, the model is trained on labeled samples using Cross Entropy Loss and Supervised Contrastive Loss\cite{khosla2020supervised}, while also trained on all samples from train set using traditional contrastive loss for even better representation. In clustering stage, the model is trained on samples with pseudo soft assignments produced by Sinkhorn-Knopp algorithm\cite{cuturi2013sinkhorn} using Cross Entropy Loss and Supervised Contrastive Loss\cite{khosla2020supervised}, while also trained on labeled samples as in warm up stage.}
  \label{fig1}
\end{figure*}

In this section, we are going to introduce Deep Contrastive Semi-supervised Clustering (DCSC), which is for discovering new intents from raw corpus. The training procedure of DCSC mainly includes two stages, warm up stage and clustering stage, which will be discussed in detail. The overall modeling procedure of DCSC is shown as Fig.\ref{fig1}.

\subsection{Warm Up Stage}
Our backbone is a language model (such as BERT\cite{devlin2018bert} and MPNet\cite{song2020mpnet}). Based on the last hidden states of backbone, we use mean pooling to get an instance vector of hidden size \begin{math}D\end{math}, and then build one more dense layer to get the final representation. For data augmentation, we use the same strategy as SimCSE\cite{gao2021simcse}, make an instance propagate through the backbone with dropout twice to get different views. Considering a batch \begin{math}X=\left \{x_{1},x_{2},\cdots,x_{N} \right \}\end{math} with batch size \begin{math}N\end{math}, we can let it through the backbone and get two output representations (views), which are \begin{math}Z=\left \{z_{1},z_{2},\cdots,z_{N} \right \}\end{math} and \begin{math}Z'=\left \{z_{N+1},z_{N+2},\cdots,z_{2N}\right \}\end{math}, and corresponding index \begin{math}I=\left \{1,2,\cdots,2N\right \}\end{math} as well.

For supervised learning on labeled samples, with labels \begin{math}Y=\left \{y_{1},y_{2},\cdots,y_{N} \right \}\end{math}, we can calculate cross entropy loss \begin{math}\mathcal{L}_{ce}^{warm up}\end{math}:
\begin{equation}
\mathcal{L}_{ce}^{warm up}=-\frac{1}{N}{\textstyle\sum_{i=1}^{N}}\mathrm{log}\frac{\mathrm{exp}\left(w_{y_{i}}\cdot z_{i}\right)}{{\textstyle\sum_{j=1}^{K}\mathrm{exp}\left(w_{j}\cdot z_{i}\right)}},
\end{equation}
where \begin{math}K\end{math} is the number of known intents, \begin{math}W=\left \{w_{1},w_{2},\cdots,w_{K} \right \}\end{math} is the classifier weights with shape \begin{math}\left(K,D\right)\end{math}. This warm up step is the same as DeepAligned\cite{zhang2021discovering} (except DeepAligned doesn't augment samples for two views), but we think such classification task will not produce an ideal representation space for clustering. Therefore, we add another Supervised Contrastive Loss\cite{khosla2020supervised} to readjust distance between any two instances according to whether they belong to the same class or not. Therefore, we can calculate supervised contrastive loss \begin{math}\mathcal{L}_{sc}^{warm up}\end{math}:
\begin{equation}
\mathcal{L}_{sc}^{warm up}={\textstyle\sum_{i=1}^{2N}}\frac{-1}{\left|P\left(i\right)\right|}{\textstyle\sum_{p\in P\left(i\right)}^{}}\mathrm{log}\frac{\mathrm{exp}\left(z_{i}\cdot z_{p}/\tau\right)}{{\textstyle\sum_{j\in A\left(i\right)}^{}}\mathrm{exp}\left(z_{i}\cdot z_{j}/\tau\right)},
\end{equation}
where \begin{math}A\left(i\right)\equiv I\setminus\left\{i\right\}\end{math}, \begin{math}P\left(i\right)\equiv\left\{p\in A\left(i\right):\tilde{y}_{p}=\tilde{y}_{i}\right\}\end{math}, and \begin{math}\tau\in\mathcal{R}^{+}\end{math}. Therefore, we have the loss for supervised learning \begin{math}\mathcal{L}_{sup}^{warm up}\end{math}:
\begin{equation}
\mathcal{L}_{sup}^{warm up}=\mathcal{L}_{ce}^{cluster}+\mathcal{L}_{sc}^{cluster}.
\end{equation}

Furthermore, we apply traditional contrastive loss on all samples, trying to improve the initial representation further from backbone before the clustering stage. We can calculate contrastive loss \begin{math}\mathcal{L}_{unsup}^{warm up}\end{math}:
\begin{equation}
\mathcal{L}_{unsup}^{warm up}=-{\textstyle\sum_{i=1}^{2N}}\mathrm{log}\frac{\mathrm{exp}\left(z_{i}\cdot z_{m\left(i\right)}/\tau\right)}{{\textstyle\sum_{j\in A\left(i\right)}^{}}\mathrm{exp}\left(z_{i}\cdot z_{j}/\tau\right)},
\end{equation}
where \begin{math}z_{m\left(i\right)}\end{math} indicates another view augmented from the same instance as \begin{math}z_{i}\end{math}. In our warm up stage, we alternately input a batch for supervised training or unsupervised training, which is like a dual-task procedure.

\subsection{Clustering Stage}
After warm up stage, we are going to initialize the weights of the cluster head at first. We extract the representations for all instances using the trained backbone, and we apply K-Means++\cite{arthur2006k} on the representations to get cluster centers $C'$ with shape \begin{math}\left(G,D\right)\end{math}, where $G$ is the ground truth number of intents (we are not going to investigate how to estimate G in this paper). Then, we use Hungarian algorithm\cite{kuhn1955hungarian} to find the optimal mapping between $W$ and $C'$, since $W$ contains a subset of intents ($K<G$), we extract the centers most likely to be the known intents from $C'$ and get \begin{math}C=\left\{c_{1},c_{2},\cdots,c_{K} \right\}\end{math} with the corresponding index as $W$. For simplicity, we can resort $C'$ as \begin{math}C'=\left\{c_{1},c_{2},\cdots,c_{K},c_{K+1},c_{K+2},\cdots,c_{G} \right\}\end{math}. The reason why we need to extract the centers of known intents will be discussed in the last paragraph of this subsection.

In clustering stage, as in warm up, we input a batch to get pairs \begin{math}Z=\left \{z_{1},z_{2},\cdots,z_{N} \right \}\end{math} and \begin{math}Z'=\left \{z_{N+1},z_{N+2},\cdots,z_{2N}\right \}\end{math}. For self-supervised clustering, we mainly refer to they way of SwAV\cite{caron2020unsupervised} training representations for images. In detail, firstly we calculate the prediction logits from cluster head for $Z$ and $Z'$ and get \begin{math}Q=\left \{q_{1},q_{2},\cdots,q_{N} \right \}\end{math} and \begin{math}Q'=\left \{q_{N+1},q_{N+2},\cdots,q_{2N}\right \}\end{math}, where:
\begin{equation}
q_{ij}=c_{j}\cdot z_{i},\forall i\in\left\{1,\cdots,2N\right\},j\in\left\{1,\cdots,G\right\}.
\end{equation}
Then we use Sinkhorn-Knopp algorithm\cite{cuturi2013sinkhorn} to get soft pseudo cluster assignments for $Q$ and $Q'$, as \begin{math}A=\left \{a_{1},a_{2},\cdots,a_{N} \right \}\end{math} and \begin{math}A'=\left \{a_{N+1},a_{N+2},\cdots,a_{2N} \right \}\end{math}, with the shape $\left(N,G\right)$. Also, we can use argmax to get the hard assignments as \begin{math}B=\left \{b_{1},b_{2},\cdots,b_{N} \right \}\end{math} and \begin{math}B'=\left \{b_{N+1},b_{N+2},\cdots,b_{2N} \right \}\end{math}, with the shape $\left(N\right)$. Sinkhorn-Knopp algorithm set the soft assignment for a instance considering not only its own logits, but also the other logits from the same batch, which can calculate the optimal distribution in a batch for all intents and avoid trivial solution. According to the soft pseudo assignments, we can calculate "swapped" cross entropy loss\cite{caron2020unsupervised} \begin{math}\mathcal{L}_{sinkhorn}^{cluster}\end{math}:
\begin{equation}
\mathcal{L}_{left}^{cluster}=-\frac{1}{N}{\textstyle\sum_{i=1}^{N}}{\textstyle\sum_{j=1}^{G}}\left(a_{\left(i+N\right)j}\cdot\mathrm{log}\frac{\mathrm{exp}\left(q_{ij}\right)}{{\textstyle\sum_{r=1}^{G}}\mathrm{exp}\left(q_{ir}\right)}\right),
\end{equation}
\begin{equation}
\mathcal{L}_{right}^{cluster}=-\frac{1}{N}{\textstyle\sum_{i=N+1}^{2N}}{\textstyle\sum_{j=1}^{G}}\left(a_{\left(i-N\right)j}\cdot\mathrm{log}\frac{\mathrm{exp}\left(q_{ij}\right)}{{\textstyle\sum_{r=1}^{G}}\mathrm{exp}\left(q_{ir}\right)}\right),
\end{equation}
\begin{equation}
\mathcal{L}_{sinkhorn}^{cluster}=\left(\mathcal{L}_{left}^{cluster}+\mathcal{L}_{right}^{cluster}\right)/2,
\end{equation}
where \begin{math}c\in C'\end{math}. To make samples belong to the same cluster closer in the representation space and get better clustering performance, we also add supervised contrastive loss according to the pseudo labels $B$ and $B'$, and get \begin{math}\mathcal{L}_{pseudo}^{cluster}\end{math}:
\begin{equation}
\mathcal{L}_{pseudo}^{cluster}={\textstyle\sum_{i=1}^{2N}}\frac{-1}{\left|H\left(i\right)\right|}{\textstyle\sum_{h\in H\left(i\right)}^{}}\mathrm{log}\frac{\mathrm{exp}\left(z_{i}\cdot z_{p}/\tau\right)}{{\textstyle\sum_{j\in A\left(i\right)}^{}}\mathrm{exp}\left(z_{i}\cdot z_{j}/\tau\right)},
\end{equation}
where \begin{math}H\left(i\right)\equiv\left\{h\in A\left(i\right):\tilde{b}_{p}=\tilde{b}_{i}\right\}\end{math}. Thus, the final loss for our deep clustering is:
\begin{equation}
\mathcal{L}_{main}^{cluster}=\mathcal{L}_{sinkhorn}^{cluster}+\mathcal{L}_{pseudo}^{cluster}.
\end{equation}

With the clustering stage discussed above, we notice that the classification accuracy on known intents decreases after several epochs, which seems like that the model "forgets" the information learned in warm up stage. This phenomenon might also reduce the clustering performance. To maintain the performance on classifying known intents, we keep the model trained on labeled instances. To let the labeled information better guide the clustering learning, we make the classifier layer and cluster layer share the same weights $C$, which is mentioned in the first paragraph of this subsection. Thus, we can calculate supervised loss \begin{math}\mathcal{L}_{sup}^{cluster}\end{math}:
\begin{equation}
\mathcal{L}_{ce}^{cluster}=-\frac{1}{N}{\textstyle\sum_{i=1}^{N}}\mathrm{log}\frac{\mathrm{exp}\left(c_{y_{i}}\cdot z_{i}\right)}{{\textstyle\sum_{j=1}^{K}\mathrm{exp}\left(c_{j}\cdot z_{i}\right)}},\forall c\in C,
\end{equation}
\begin{equation}
\mathcal{L}_{sc}^{cluster}={\textstyle\sum_{i=1}^{2N}}\frac{-1}{\left|P\left(i\right)\right|}{\textstyle\sum_{p\in P\left(i\right)}^{}}\mathrm{log}\frac{\mathrm{exp}\left(z_{i}\cdot z_{p}/\tau\right)}{{\textstyle\sum_{j\in A\left(i\right)}^{}}\mathrm{exp}\left(z_{i}\cdot z_{j}/\tau\right)},
\end{equation}
\begin{equation}
\mathcal{L}_{sup}^{cluster}=\mathcal{L}_{ce}^{cluster}+\mathcal{L}_{sc}^{cluster}.
\end{equation}
In the clustering stage, as in warm up, we also alternately input a batch for supervised learning or cluster learning. Therefore, we can get the overall training process as shown in Fig.\ref{fig1}.

\section{Experiments}
In this section, we introduce the details of our experiments, and discuss our model performance specifically.

\subsection{Datasets}
We conduct experiments on two public datasets consist of user queries and labeled intents. Details are shown in Table \ref{table1}.

\textbf{Banking.} It provides user queries and labeled intents from banking domain for text classification or text clustering, with totally 13083 samples and 77 types of intents\cite{casanueva2020efficient}.

\textbf{Clinc.} It contains 22500 samples of user queries in total and 150 unique labeled intents, which can be used for text classification or text clustering as well\cite{larson2019evaluation}.

\subsection{Baselines}
We choose currently popular methods for discovering new intents using semi-supervised clustering, including DeepAligned\cite{zhang2021discovering}, and SCL\cite{shen2021semi}. We directly report the results of these baselines from their papers if the results are available, otherwise we run the official code with current experiment settings and make the report.

\subsection{Experiment Settings}
We keep the same evaluation settings as in DeepAligned\cite{zhang2021discovering} for intuitive comparison. Specifically, we keep the same data split as DeepAligned for training set, validation set, and test set. To simulate the scenario as discovering new intents from raw corpus, we randomly select a certain percentage of intents as known ($25\%$, $50\%$, and $75\%$ in our cases), and then randomly select $10\%$ queries of known intents as labeled instances to get a new labeled subset, and treat the remaining samples as unlabeled ones. The models can be trained on the unlabeled training set and the labeled subset, and will be evaluated for clustering performance on test set.

For our method, we train DCSC in warm up stage and clustering stage for both 100 epochs. We set batch size 512 only for cluster training, and 128 for other cases. To optimize the net works, we use AdamW optimizer\cite{loshchilov2017decoupled} with learning rate 0.00005 and decaying rate 0.01. Besides, as DeepAligned\cite{zhang2021discovering} does, we freeze the weights of embedding layer and all transformer layer except the last one during training, which will not reduce the performance but will greatly improve the efficiency. For fair comparison with baselines, we have tested BERT\footnote[1]{The official pretrained "bert-base-uncased"\cite{devlin2018bert} available on Hugging Face\cite{wolf2020transformers}} and MPNet\footnote[2]{MPNet\cite{song2020mpnet} that is further pretrained for better sentence embedding\cite{reimers2019sentence} "sentence-transformers/paraphrase-mpnet-base-v2" available on Hugging Face\cite{wolf2020transformers}} as backbone of our model. After training, we extract the representation of sentences from test set and conduct K-Means++ to predict cluster assignments for final evaluation.

\subsection{Evaluation Metrics}
To evaluate the performance of models, we use Accuracy (ACC), Adjusted Rand Index (ARI), Normalized Mutual Information (NMI), which are commonly used to evaluate clustering performance.

\begin{table}
  \centering
  \caption{Statistics of Banking and Clinc, where "Classes" indicates the number of unique intents, and "Training", "Validation", "Test" indicate the number of instances in the corresponding set.}
  \label{table1}
  \begin{tabular}{ccccc}
    \toprule
    Dataset&Classes&Training&Validation&Test\\
    \midrule
    Banking&77&9003&1000&3080\\
    Clinc&150&18000&2250&2250\\
    \bottomrule
  \end{tabular}
\end{table}

\subsection{Main Results}
We have summarized the clustering results from different methods in Table \ref{table2}, and DCSC has achieved the best results across all settings and datasets, which indicates its robustness and accuracy in various scenarios.

\begin{table*}[ht]
  \centering
  \caption{Clustering results of different experiment settings and datasets, where $25\%$, $50\%$, and $75\%$ indicate the fraction of known intents. $\dagger$ indicates the result we have actually run, $\ddagger$ indicates the result reported in SCL\cite{shen2021semi}, otherwise the result is reported from its own paper.}
  \label{table2}
  \begin{tabular}{cccccccccccc}
    \toprule
    \multicolumn{1}{c}{}&\multicolumn{1}{c}{}&\multicolumn{1}{c}{}&\multicolumn{3}{c}{$25\%$}&\multicolumn{3}{c}{$50\%$}&\multicolumn{3}{c}{$75\%$}\\
    \cmidrule(lr){4-6}
    \cmidrule(lr){7-9}
    \cmidrule(lr){10-12}
    Dataset&Backbone&Model&ACC&ARI&NMI&ACC&ARI&NMI&ACC&ARI&NMI\\
    \midrule
    \multirow{4}{*}{Banking}
    &BERT&DeepAligned&$49.51^{\dagger}$&$37.29^{\dagger}$&$70.26^{\dagger}$&$59.44^{\ddagger}$&$47.07^{\ddagger}$&$76.14^{\ddagger}$&64.90&53.64&79.56\\
    &MPNet&SCL&58.73&47.47&76.79&67.28&55.50&80.25&76.55&65.43&85.04\\
    &BERT&DCSC&60.15&49.75&78.18&68.30&56.94&81.19&75.18&64.55&84.65\\
    &MPNet&DCSC&\textbf{68.85}&\textbf{58.41}&\textbf{82.26}&\textbf{74.05}&\textbf{63.03}&\textbf{84.56}&\textbf{77.54}&\textbf{67.92}&\textbf{86.59}\\
    \midrule
    \multirow{3}{*}{Clinc}
    &BERT&DeepAligned&$75.20^{\ddagger}$&$65.36^{\ddagger}$&$89.12^{\ddagger}$&$80.70^{\ddagger}$&$75.26^{\ddagger}$&$91.50^{\ddagger}$&86.49&79.75&93.89\\
    &MPNet&SCL&71.23&62.02&88.30&78.36&70.71&91.38&86.91&81.64&94.75\\
    &BERT&DCSC&\textbf{79.89}&\textbf{72.68}&\textbf{91.70}&\textbf{84.57}&\textbf{78.82}&\textbf{93.75}&\textbf{89.70}&\textbf{84.41}&\textbf{95.28}\\
    \bottomrule
  \end{tabular}
\end{table*}

\textbf{Effect of cluster learning.}
First of all, $\mathrm{DCSC_{BERT}}$ has a better clustering performance than DeepAligned across all situations, and it also has outperformed SCL (with a better backbone) in most cases. SCL directly train the backbone on labeled queries with contrastive learning, while it doesn't make a self-supervised cluster training to optimize the representation space further as DeepAligned and DCSC does. In the settings of $25\%$ and $50\%$ known intents for Clinc dataset, the results of SCL is worse than DeepAligned, which indicates current method is not robust enough and there's a large potential for improvement considering MPNet\footnotemark[2] should be better on extracting sentence embeddings. Even compared with DeepAligned, DCSC is more efficient and accurate at the cluster learning stage. DeepAligned use K-Means to update pseudo labels, so it requires encoding all training instances additionally and apply K-Means for clustering after every epoch. DCSC doesn't predict pseudo labels globally, it assign pseudo labels simultaneous when given a training batch. Furthermore, DCSC jointly optimize the instance representation and cluster assignments, which can better guide the clustering procedure. Besides, though $\mathrm{DCSC_{BERT}}$ has improved the results a lot, $\mathrm{DCSC_{MPNet}}$ can achieve even better results with MPNet as backbone. Thus, better backbone or better initial sentence embedding, is still an improvement method worth trying.

\textbf{Effect of contrastive learning.}
Both DCSC and SCL use contrastive learning for better representation, although through different methods. DeepAligned mainly relies on classification loss to optimize representation, which is weak since it lacks margin constrain of hidden space for clustering based on distance.

\subsection{Ablation Study}
In this subsection, we analyze the effect of our model improvements through ablation studies.

\begin{table}[ht]
  \centering
  \caption{Ablation study of DCSC with BERT as backbone, where $\dagger$ indicates without supervised training in clustering stage.}
  \label{table2}
  \begin{tabular}{cccccc}
    \toprule
    Dataset&Fraction&Method&ACC&ARI&NMI\\
    \midrule
    \multirow{6}{*}{Banking}
    &\multirow{2}{*}{$25\%$}
    &$\mathrm{DCSC^{\dagger}}$&57.32&48.43&77.81\\
    &&DCSC&\textbf{60.15}&\textbf{49.75}&\textbf{78.18}\\
    \cmidrule(lr){2-6}
    &\multirow{2}{*}{$50\%$}
    &$\mathrm{DCSC^{\dagger}}$&62.80&53.54&80.48\\
    &&DCSC&\textbf{68.30}&\textbf{56.94}&\textbf{81.19}\\
    \cmidrule(lr){2-6}
    &\multirow{2}{*}{$75\%$}
    &$\mathrm{DCSC^{\dagger}}$&65.36&56.65&82.13\\
    &&DCSC&\textbf{75.18}&\textbf{64.55}&\textbf{84.65}\\
    \midrule
    \multirow{6}{*}{Clinc}
    &\multirow{2}{*}{$25\%$}
    &$\mathrm{DCSC^{\dagger}}$&78.22&\textbf{72.84}&\textbf{92.77}\\
    &&DCSC&\textbf{79.89}&72.68&91.70\\
    \cmidrule(lr){2-6}
    &\multirow{2}{*}{$50\%$}
    &$\mathrm{DCSC^{\dagger}}$&81.85&76.63&\textbf{93.81}\\
    &&DCSC&\textbf{84.57}&\textbf{78.82}&93.75\\
    \cmidrule(lr){2-6}
    &\multirow{2}{*}{$75\%$}
    &$\mathrm{DCSC^{\dagger}}$&83.42&78.46&94.31\\
    &&DCSC&\textbf{89.70}&\textbf{84.81}&\textbf{95.28}\\
    \bottomrule
  \end{tabular}
\end{table}

\textbf{Supervised training in clustering stage.} After comparing $\mathrm{DCSC^{\dagger}}$ and DCSC, we can figure out that the more intents are known, the more the clustering performance will decreases. This is because the model will learn complete supervised information during warm up stage in the setting of $50\%$ and $75\%$ known intents, thus it will drop more information in clustering stage without the guide of classification label.

\section{Conclusion}
In this paper, we propose Deep Contrastive Semi-supervised Clustering (DCSC), which is for discovering new intents from raw user queries. DCSC is trained through a two-stage dual-task process, to fully utilize the limited supervised information and improve the representation space with contrastive learning. Furthermore, DCSC builds a deep-learning-based clustering approach as a semi-supervised tasks, which jointly optimize the clustering and the representation to improve the final performance. We compare our model with other methods through the experiments on two public datasets, and DCSC has achieved the best results across all experiment settings and datasets, indicating that the improvements we've made can greatly improve the robustness and accuracy on text clustering.

\bibliographystyle{ACM-Reference-Format}
\bibliography{sample-base}


\begin{thebibliography}{22}

%%% the Web address.

\ifx \showCODEN    \undefined \def \showCODEN     #1{\unskip}     \fi
\ifx \showDOI      \undefined \def \showDOI       #1{#1}\fi
\ifx \showISBNx    \undefined \def \showISBNx     #1{\unskip}     \fi
\ifx \showISBNxiii \undefined \def \showISBNxiii  #1{\unskip}     \fi
\ifx \showISSN     \undefined \def \showISSN      #1{\unskip}     \fi
\ifx \showLCCN     \undefined \def \showLCCN      #1{\unskip}     \fi
\ifx \shownote     \undefined \def \shownote      #1{#1}          \fi
\ifx \showarticletitle \undefined \def \showarticletitle #1{#1}   \fi
\ifx \showURL      \undefined \def \showURL       {\relax}        \fi
% invisible to TeX
\providecommand\bibfield[2]{#2}
\providecommand\bibinfo[2]{#2}
\providecommand\natexlab[1]{#1}
\providecommand\showeprint[2][]{arXiv:#2}

\bibitem[Arthur and Vassilvitskii(2006)]%
        {arthur2006k}
\bibfield{author}{\bibinfo{person}{David Arthur} {and} \bibinfo{person}{Sergei
  Vassilvitskii}.} \bibinfo{year}{2006}\natexlab{}.
\newblock \bibinfo{booktitle}{\emph{k-means++: The advantages of careful
  seeding}}.
\newblock \bibinfo{type}{{T}echnical {R}eport}.
  \bibinfo{institution}{Stanford}.
\newblock


\bibitem[Campello et~al\mbox{.}(2013)]%
        {campello2013density}
\bibfield{author}{\bibinfo{person}{Ricardo~JGB Campello},
  \bibinfo{person}{Davoud Moulavi}, {and} \bibinfo{person}{J{\"o}rg Sander}.}
  \bibinfo{year}{2013}\natexlab{}.
\newblock \showarticletitle{Density-based clustering based on hierarchical
  density estimates}. In \bibinfo{booktitle}{\emph{Pacific-Asia conference on
  knowledge discovery and data mining}}. Springer, \bibinfo{pages}{160--172}.
\newblock


\bibitem[Caron et~al\mbox{.}(2018)]%
        {caron2018deep}
\bibfield{author}{\bibinfo{person}{Mathilde Caron}, \bibinfo{person}{Piotr
  Bojanowski}, \bibinfo{person}{Armand Joulin}, {and} \bibinfo{person}{Matthijs
  Douze}.} \bibinfo{year}{2018}\natexlab{}.
\newblock \showarticletitle{Deep clustering for unsupervised learning of visual
  features}. In \bibinfo{booktitle}{\emph{Proceedings of the European
  Conference on Computer Vision (ECCV)}}. \bibinfo{pages}{132--149}.
\newblock


\bibitem[Caron et~al\mbox{.}(2020)]%
        {caron2020unsupervised}
\bibfield{author}{\bibinfo{person}{Mathilde Caron}, \bibinfo{person}{Ishan
  Misra}, \bibinfo{person}{Julien Mairal}, \bibinfo{person}{Priya Goyal},
  \bibinfo{person}{Piotr Bojanowski}, {and} \bibinfo{person}{Armand Joulin}.}
  \bibinfo{year}{2020}\natexlab{}.
\newblock \showarticletitle{Unsupervised learning of visual features by
  contrasting cluster assignments}.
\newblock \bibinfo{journal}{\emph{arXiv preprint arXiv:2006.09882}}
  (\bibinfo{year}{2020}).
\newblock


\bibitem[Casanueva et~al\mbox{.}(2020)]%
        {casanueva2020efficient}
\bibfield{author}{\bibinfo{person}{Inigo Casanueva}, \bibinfo{person}{Tadas
  Tem{\v{c}}inas}, \bibinfo{person}{Daniela Gerz}, \bibinfo{person}{Matthew
  Henderson}, {and} \bibinfo{person}{Ivan Vuli{\'c}}.}
  \bibinfo{year}{2020}\natexlab{}.
\newblock \showarticletitle{Efficient intent detection with dual sentence
  encoders}.
\newblock \bibinfo{journal}{\emph{arXiv preprint arXiv:2003.04807}}
  (\bibinfo{year}{2020}).
\newblock


\bibitem[Chen et~al\mbox{.}(2020)]%
        {chen2020simple}
\bibfield{author}{\bibinfo{person}{Ting Chen}, \bibinfo{person}{Simon
  Kornblith}, \bibinfo{person}{Mohammad Norouzi}, {and}
  \bibinfo{person}{Geoffrey Hinton}.} \bibinfo{year}{2020}\natexlab{}.
\newblock \showarticletitle{A simple framework for contrastive learning of
  visual representations}. In \bibinfo{booktitle}{\emph{International
  conference on machine learning}}. PMLR, \bibinfo{pages}{1597--1607}.
\newblock


\bibitem[Cuturi(2013)]%
        {cuturi2013sinkhorn}
\bibfield{author}{\bibinfo{person}{Marco Cuturi}.}
  \bibinfo{year}{2013}\natexlab{}.
\newblock \showarticletitle{Sinkhorn distances: Lightspeed computation of
  optimal transport}.
\newblock \bibinfo{journal}{\emph{Advances in neural information processing
  systems}}  \bibinfo{volume}{26} (\bibinfo{year}{2013}),
  \bibinfo{pages}{2292--2300}.
\newblock


\bibitem[Devlin et~al\mbox{.}(2018)]%
        {devlin2018bert}
\bibfield{author}{\bibinfo{person}{Jacob Devlin}, \bibinfo{person}{Ming-Wei
  Chang}, \bibinfo{person}{Kenton Lee}, {and} \bibinfo{person}{Kristina
  Toutanova}.} \bibinfo{year}{2018}\natexlab{}.
\newblock \showarticletitle{Bert: Pre-training of deep bidirectional
  transformers for language understanding}.
\newblock \bibinfo{journal}{\emph{arXiv preprint arXiv:1810.04805}}
  (\bibinfo{year}{2018}).
\newblock


\bibitem[Gao et~al\mbox{.}(2021)]%
        {gao2021simcse}
\bibfield{author}{\bibinfo{person}{Tianyu Gao}, \bibinfo{person}{Xingcheng
  Yao}, {and} \bibinfo{person}{Danqi Chen}.} \bibinfo{year}{2021}\natexlab{}.
\newblock \showarticletitle{SimCSE: Simple Contrastive Learning of Sentence
  Embeddings}.
\newblock \bibinfo{journal}{\emph{arXiv preprint arXiv:2104.08821}}
  (\bibinfo{year}{2021}).
\newblock


\bibitem[Khosla et~al\mbox{.}(2020)]%
        {khosla2020supervised}
\bibfield{author}{\bibinfo{person}{Prannay Khosla}, \bibinfo{person}{Piotr
  Teterwak}, \bibinfo{person}{Chen Wang}, \bibinfo{person}{Aaron Sarna},
  \bibinfo{person}{Yonglong Tian}, \bibinfo{person}{Phillip Isola},
  \bibinfo{person}{Aaron Maschinot}, \bibinfo{person}{Ce Liu}, {and}
  \bibinfo{person}{Dilip Krishnan}.} \bibinfo{year}{2020}\natexlab{}.
\newblock \showarticletitle{Supervised contrastive learning}.
\newblock \bibinfo{journal}{\emph{arXiv preprint arXiv:2004.11362}}
  (\bibinfo{year}{2020}).
\newblock


\bibitem[Kuhn(1955)]%
        {kuhn1955hungarian}
\bibfield{author}{\bibinfo{person}{Harold~W Kuhn}.}
  \bibinfo{year}{1955}\natexlab{}.
\newblock \showarticletitle{The Hungarian method for the assignment problem}.
\newblock \bibinfo{journal}{\emph{Naval research logistics quarterly}}
  \bibinfo{volume}{2}, \bibinfo{number}{1-2} (\bibinfo{year}{1955}),
  \bibinfo{pages}{83--97}.
\newblock


\bibitem[Larson et~al\mbox{.}(2019)]%
        {larson2019evaluation}
\bibfield{author}{\bibinfo{person}{Stefan Larson}, \bibinfo{person}{Anish
  Mahendran}, \bibinfo{person}{Joseph~J Peper}, \bibinfo{person}{Christopher
  Clarke}, \bibinfo{person}{Andrew Lee}, \bibinfo{person}{Parker Hill},
  \bibinfo{person}{Jonathan~K Kummerfeld}, \bibinfo{person}{Kevin Leach},
  \bibinfo{person}{Michael~A Laurenzano}, \bibinfo{person}{Lingjia Tang},
  {et~al\mbox{.}}} \bibinfo{year}{2019}\natexlab{}.
\newblock \showarticletitle{An evaluation dataset for intent classification and
  out-of-scope prediction}.
\newblock \bibinfo{journal}{\emph{arXiv preprint arXiv:1909.02027}}
  (\bibinfo{year}{2019}).
\newblock


\bibitem[Lin et~al\mbox{.}(2020)]%
        {lin2020discovering}
\bibfield{author}{\bibinfo{person}{Ting-En Lin}, \bibinfo{person}{Hua Xu},
  {and} \bibinfo{person}{Hanlei Zhang}.} \bibinfo{year}{2020}\natexlab{}.
\newblock \showarticletitle{Discovering new intents via constrained deep
  adaptive clustering with cluster refinement}. In
  \bibinfo{booktitle}{\emph{Proceedings of the AAAI Conference on Artificial
  Intelligence}}, Vol.~\bibinfo{volume}{34}. \bibinfo{pages}{8360--8367}.
\newblock


\bibitem[Loshchilov and Hutter(2017)]%
        {loshchilov2017decoupled}
\bibfield{author}{\bibinfo{person}{Ilya Loshchilov} {and}
  \bibinfo{person}{Frank Hutter}.} \bibinfo{year}{2017}\natexlab{}.
\newblock \showarticletitle{Decoupled weight decay regularization}.
\newblock \bibinfo{journal}{\emph{arXiv preprint arXiv:1711.05101}}
  (\bibinfo{year}{2017}).
\newblock


\bibitem[Reimers and Gurevych(2019)]%
        {reimers2019sentence}
\bibfield{author}{\bibinfo{person}{Nils Reimers} {and} \bibinfo{person}{Iryna
  Gurevych}.} \bibinfo{year}{2019}\natexlab{}.
\newblock \showarticletitle{Sentence-bert: Sentence embeddings using siamese
  bert-networks}.
\newblock \bibinfo{journal}{\emph{arXiv preprint arXiv:1908.10084}}
  (\bibinfo{year}{2019}).
\newblock


\bibitem[Shen et~al\mbox{.}(2021)]%
        {shen2021semi}
\bibfield{author}{\bibinfo{person}{Xiang Shen}, \bibinfo{person}{Yinge Sun},
  \bibinfo{person}{Yao Zhang}, {and} \bibinfo{person}{Mani Najmabadi}.}
  \bibinfo{year}{2021}\natexlab{}.
\newblock \showarticletitle{Semi-supervised Intent Discovery with Contrastive
  Learning}. In \bibinfo{booktitle}{\emph{Proceedings of the 3rd Workshop on
  Natural Language Processing for Conversational AI}}.
  \bibinfo{pages}{120--129}.
\newblock


\bibitem[Song et~al\mbox{.}(2020)]%
        {song2020mpnet}
\bibfield{author}{\bibinfo{person}{Kaitao Song}, \bibinfo{person}{Xu Tan},
  \bibinfo{person}{Tao Qin}, \bibinfo{person}{Jianfeng Lu}, {and}
  \bibinfo{person}{Tie-Yan Liu}.} \bibinfo{year}{2020}\natexlab{}.
\newblock \showarticletitle{Mpnet: Masked and permuted pre-training for
  language understanding}.
\newblock \bibinfo{journal}{\emph{arXiv preprint arXiv:2004.09297}}
  (\bibinfo{year}{2020}).
\newblock


\bibitem[Wolf et~al\mbox{.}(2020)]%
        {wolf2020transformers}
\bibfield{author}{\bibinfo{person}{Thomas Wolf}, \bibinfo{person}{Julien
  Chaumond}, \bibinfo{person}{Lysandre Debut}, \bibinfo{person}{Victor Sanh},
  \bibinfo{person}{Clement Delangue}, \bibinfo{person}{Anthony Moi},
  \bibinfo{person}{Pierric Cistac}, \bibinfo{person}{Morgan Funtowicz},
  \bibinfo{person}{Joe Davison}, \bibinfo{person}{Sam Shleifer},
  {et~al\mbox{.}}} \bibinfo{year}{2020}\natexlab{}.
\newblock \showarticletitle{Transformers: State-of-the-art natural language
  processing}. In \bibinfo{booktitle}{\emph{Proceedings of the 2020 Conference
  on Empirical Methods in Natural Language Processing: System Demonstrations}}.
  \bibinfo{pages}{38--45}.
\newblock


\bibitem[Xie et~al\mbox{.}(2016)]%
        {xie2016unsupervised}
\bibfield{author}{\bibinfo{person}{Junyuan Xie}, \bibinfo{person}{Ross
  Girshick}, {and} \bibinfo{person}{Ali Farhadi}.}
  \bibinfo{year}{2016}\natexlab{}.
\newblock \showarticletitle{Unsupervised deep embedding for clustering
  analysis}. In \bibinfo{booktitle}{\emph{International conference on machine
  learning}}. PMLR, \bibinfo{pages}{478--487}.
\newblock


\bibitem[Yang et~al\mbox{.}(2017)]%
        {yang2017towards}
\bibfield{author}{\bibinfo{person}{Bo Yang}, \bibinfo{person}{Xiao Fu},
  \bibinfo{person}{Nicholas~D Sidiropoulos}, {and} \bibinfo{person}{Mingyi
  Hong}.} \bibinfo{year}{2017}\natexlab{}.
\newblock \showarticletitle{Towards k-means-friendly spaces: Simultaneous deep
  learning and clustering}. In \bibinfo{booktitle}{\emph{international
  conference on machine learning}}. PMLR, \bibinfo{pages}{3861--3870}.
\newblock


\bibitem[Zhang et~al\mbox{.}(2021a)]%
        {zhang2021supporting}
\bibfield{author}{\bibinfo{person}{Dejiao Zhang}, \bibinfo{person}{Feng Nan},
  \bibinfo{person}{Xiaokai Wei}, \bibinfo{person}{Shangwen Li},
  \bibinfo{person}{Henghui Zhu}, \bibinfo{person}{Kathleen McKeown},
  \bibinfo{person}{Ramesh Nallapati}, \bibinfo{person}{Andrew Arnold}, {and}
  \bibinfo{person}{Bing Xiang}.} \bibinfo{year}{2021}\natexlab{a}.
\newblock \showarticletitle{Supporting Clustering with Contrastive Learning}.
\newblock \bibinfo{journal}{\emph{arXiv preprint arXiv:2103.12953}}
  (\bibinfo{year}{2021}).
\newblock


\bibitem[Zhang et~al\mbox{.}(2021b)]%
        {zhang2021discovering}
\bibfield{author}{\bibinfo{person}{Hanlei Zhang}, \bibinfo{person}{Hua Xu},
  \bibinfo{person}{Ting-En Lin}, {and} \bibinfo{person}{Rui Lyu}.}
  \bibinfo{year}{2021}\natexlab{b}.
\newblock \showarticletitle{Discovering new intents with deep aligned
  clustering}. In \bibinfo{booktitle}{\emph{Proceedings of the AAAI Conference
  on Artificial Intelligence}}, Vol.~\bibinfo{volume}{35}.
  \bibinfo{pages}{14365--14373}.
\newblock


\end{thebibliography}

% \appendix









\end{document}